\documentclass[11pt]{amsart}
\usepackage{geometry}                
\usepackage{graphicx}
\usepackage{amssymb}
\usepackage{epstopdf}
\title{Models, networks and algorithmic complexity}
\author{G. Ruffini - Dec 2016 --- Starlab Technical Note, TN00339 (v0.9)}

\newcommand{\U}{\mathcal U}

\newcommand{\inputspace}{\mathcal X}
\newcommand{\handspace}{\mathcal H}
 
\newtheorem{definition}{Definition} 
\newtheorem{theorem}{Theorem} 
\newtheorem{conjecture}{Conjecture}

\newtheorem{model}{Model}
\usepackage{listings}             

\begin{document}
\lstset{language=Python}          

\maketitle
\tableofcontents
\section*{Abstract}
I  aim to show that models, classification or generating functions, invariances and datasets are algorithmically equivalent concepts once properly defined, and provide some concrete examples of them. I then  show that a)  neural networks  (NNs) of different kinds can be seen to implement models, b) that perturbations of inputs and nodes in  NNs  trained to optimally implement simple models propagate strongly,  c) that there is a framework in which recurrent, deep  and shallow networks can be seen to fall into a descriptive power hierarchy in agreement with notions from the theory of recursive functions. 
The motivation for these definitions and following analysis lies in the context of cognitive neuroscience, and in particular in   \cite{Ruffini:2016ac}, where the concept of model is used extensively, as is the concept of algorithmic complexity.  


\clearpage

\section{Models}

 Let us first define  formally the notion of model  as in \cite{Ruffini:2016ac} in the context of computation: 
\begin{definition}
The optimal model or k-model of a dataset is the shortest program that generates (or, equivalently, compresses) the dataset efficiently (i.e., succinctly in the Kolmogorov sense).
\end{definition}
A {\em model} is any  program that generates the dataset, optimal or not. In general, we don't have access to optimal models. Some examples of models are:
\begin{itemize}
\item A Lempel-Ziv-Welch (LZW) compressed version of a file is an implementation of a model of the data in the file (running on the programming environment that can carry out the decompression). The implementation may be a poor one if the data  does not contain regularities  that are in the form of substring repetition (e.g., as in the digits of $\pi$, which can be generated by a simple algorithm but without repetition regularities).
\item A program that provided with initial condition inputs, generates dynamical data   for some physical system (e.g., positions and velocities of particles in a gas).
\item Any physics model encoded in equations (e.g., Maxwell's equations) and associated methods (e.g., calculus) which can procedurally be computed given some initial/boundary conditions to generate data.
\end{itemize}
The following, as we will discuss, are also equivalent to models:
\begin{itemize}
\item A pattern recognition program. E.g., a feedforward neural network (NN) trained to recognize images of hands, whether it is shallow (one layer, SNN) or deep (multiple layers, DNN). Such networks essentially provide an implementation to compute a function.  
\item A recurrent NN (RNN) classifying inputs by going into an attractor (a Hopfield network). 
\item An LSTM network trained to classify sequences into labels, for example for speech identification of a given word. Again, there are invariances (who says the words, or how) encoded by the network, and again we can think of the network as encoding the model for a spoken word.  
\end{itemize}
In the human brain and in machine learning we talk about neural networks, and, in general of recurrent neural networks.  For example,  learned sequences of motor activity are encoded in many vertebrate
brains by ``complex spatio-temporal patterns of neural activity'' \cite{Hahnloser:2002aa}. 
%
%
\\

We also point out that in general the Kolmogorov Complexity of a string is uncomputable due to the halting problem of Turing machines \cite{Cover:2006aa}. However, there are ways to deal with this limitation in practice. For instance, if we limit computation to programming using  {\em primitive recursive functions (PR)}, which are described by programs that include {\tt for-loops} but not {\tt while loops}\footnote{See the BlooP and FlooP languages as discussed in \cite{Hofstadter:1979aa}}, then all such programs halt. Algorithmic complexity limited to PR is therefore computable. For example, feedforward networks  are PR, and one may in practice seek to find the simplest (shortest as a program) such feedforward NN that computes a PR function.  Of course, it may be that if one uses a more general (complete) language, a shorter program can be produced for the same PR function. 

\section{Compressing a disordered stacked dataset}
In the discussion below we will refer often to  binary (two class) classification of images to make the discussion concrete, although the reasoning is more general.  We will use the symbols $\mathbb B=\{0,1\}$  (a set with two elements) and   $\inputspace \equiv \mathbb B^{n} $ for image space (with $n$ proportional to the number of pixels in an image).   A (digital) {\em class function} is a map from $f(x): \inputspace  \rightarrow \mathbb B$ (e.g., from images to binary numbers, as in binary classification of images).   \\

The discussion is entirely framed in finite, discrete  sets. 
We  restrict the discussion throughout to discrete images, both in pixel and in digitization and function values.  Image space is thus large but finite. If we consider 24 bit/pixel images of 1000 by 1000 pixels, with size $ |\inputspace | = 2^n$, with $n=24\cdot 10 ^6$. This is a large number, but the set of binary class functions $f(x) \in \mathcal F: \inputspace  \rightarrow \mathbb B$ is much larger\footnote{We will use $\log$ to refer to $\log_{2}$ througout.}, $\log |\mathcal F| =  |\inputspace | $.\\

In practice, we can assume that the number of possible images is much smaller, however, because many images are never encountered. \\


To be  concrete, here we will refer to a {\bf stacked dataset $D_{f}$:} the dataset generated from all hand images, which we imagine as a random stack of  images $x \in \inputspace$ (a datacube).  Thus, the dataset is generated using the same rule (``hand'', which we imagine as a program that takes some inputs and generates an image of a hand)  with  varying parameters $\theta \in \mathcal P= \mathbb B^{m}$ at each iteration plus a varying background for each image, which we assume belongs to a space of possible backgrounds $\mathcal B$.  
We call the set of ``hand'' images $  \handspace \subset   \mathcal P \times  \mathcal B$ (not all combinations of background and hand realizations are possible, they have to fit). It is a finite set of size $|  \handspace| < |\mathcal P| \times | \mathcal B | $.  Although this type of dataset is not the most general, it is fitting for the discussion of neural networks below. We will also assume that the stack is disordered, to make compression more difficult\footnote{However, it is possible that the optimal compressing program will identify the underlying group structure of image generation and use that to order the images and then compress them better. This concept has been proposed by Barbour as the mechanism behind the emergence for the parameter of time, for instance, if we imagine the images to be snapshots of a dynamical system \cite{Barbour:1999aa}.}. \\

The stacked dataset, a string of bits, has length $
l(D_{f})=|\handspace|\, \log |\inputspace|
$. We assume throughout that 
$$
|\inputspace| >>|\handspace|>>1.
$$

\section{Models, class function and the generating function}

Neural networks encode functions. Eg., a feedforward network maps inputs to outputs.   In what sense is a neural network or a function  a {\em model} of  the dataset? 
As  defined, models are mathematical objects (programs) which unpack into a dataset.  \\ 

{\bf From optimal model to generating function to class function:} Let us first see how a k-model can be used to define a classification function. To be  concrete, let us imagine the staked dataset generated from all the images  of a rotating and flexing hand we can generate with a few parameters (rotations, shape changes), a large, random stack of each image ($x \in \inputspace$). Since we generated the data simply, we know we can compress it into a program in which part of the program  describes what is invariant (generating function), and what is changing (shape, perspective).  More formally, the program can be thought as consisting of three parts. One part encodes what a hand is in algorithmic terms through an image  {\bf  generating function} of some parameters  $x=g(\theta)$, with $\theta \in \mathcal P$: 
$$
g(\theta): \mathcal P \longrightarrow \inputspace
.
$$
The second part is  a sequence $\theta_{i}$  where the hand model parameters are changed and has no structure. The last part describes the background, which we assume also  has no structure.   The optimal model (a compressing program) must thus encode a hand image generating function $x_{\theta}=g(\theta)$. A suboptimal model need not do this (e.g., in the worse case it may say ``print the following bits in the dataset: ..... '').\\

 If the model is optimal, the first part will be the algorithmically simplest program that encodes the generating function, and presumably unique.  To find it we could for example carry out all possible reordering of images in the dataset, and compare the models that our compressor produces for each case. Part of the model will be invariant to such operations---the one corresponding to the generating function.  \\
 
 The make the example a bit more concrete, the model may look like this: 
 \begin{quote}
 {\footnotesize
\begin{lstlisting}%[frame=single]  % Start your code-block
%P = [set of parameters]
%B= [set of backgrounds]

def Model:

    def HandImageCreator(theta)   
        # Inputs parameter theta which defines position, 
        # rotation, shape or other hand features.
	# Generates image of hand with white background
	# Provides a list of the background pixel IDs
	 [ ... ]
	return image, backgroundPixels 
   
   # we need a list of thetas in some order
   P= CreateParameterSet [...]
   
   for each theta in P: 
       handImage, backgroundPixels = HandImageCreator(theta)
       
       B= CreatePossibleBackgroundsList(backgroundPixels)
       
       for b in B:    	   
       		handImage.addBackground(b)		
            	place handImage in Stack
	
   return Stack
\end{lstlisting}
}
\end{quote}
We could actually create a finite listing of all possible parameter-background combinations and simply iterate over them or over a permutation of this list (this model is PR). \\ 

An useful concept to specify  the generating function is that of the {\bf Kolmogorov Structure function}, $\Phi_k$ of the dataset \cite{Cover:2006aa}.  This function is  $\Phi_k(D_{f})= \log |S|$ where $S$ is the smallest set that  1) contains the dataset as an element and 2) can be described with no more than $k$ bits.  The function equals the length of the dataset for $k=0$ ($\Phi_0(D_{f})=l$)  and 0 when the $k = \mathcal K [D_{f}]$ ($\Phi_{\mathcal K [D_{f}]}(D_{f})= 0$), since there is only one set that contains $D_{f}$). As we add more bits, we  constrain further the size of $S$, which will decrease rapidly while regularities are encoded. At some point, the rate of decrease of $\log |S|$ will be of 1 per bit, which means we are then encoding the incompressible part of the dataset---in this case constraining the random list of parameters and background. Thus, the program at the critical length $k^{*}$ \cite{Cover:2006aa}  is the one that makes use of the generating function, and may look like this ``Iterate over all possible images and keep those which have maximal correlation with the output of the generating function for some parameter $\theta$ with due care for masking. Create an element per permutation of this stack.'' Here we assume that the generating function provides images of a hand and a mask for the white background---a list of background pixels to compute the correlation with the masked image.  
The set $S^*$ defined using $k^{*}$ bits is called the ``algorithmic sufficient statistic for the dataset'', and satisfies $\mathcal K(x)+c=\mathcal K(S^{*})+\log |S^{*}| = k^{*}+\log |S^{*}|$. \\


From the generating function we can construct a classifier of hands (class function). We can write a program that given an image $x$ scans over all allowed parameter vales  (including backgrounds $b$) and outputs 1 if there exits a value such that $g(\theta,b)=x$, or zero otherwise. We can  also carry out the search with a fixed $b_{0}$ by allowing some error, $|| g(\theta, b_{0}) - x || < \epsilon$.\\

{\bf From class function to model:}  To illustrate now how a  classification or class function can be used to create  a model, let us begin with a simple type, a function defined by a feedforward network such as the ones used for image classification. Such a network encodes a function $f(x)$  that when input an image of hand $x$   outputs a value of 1, and 0 otherwise,
$$
f(x): \inputspace \longrightarrow \mathbb B.
$$ Geometrically,  this function is an invariant over the manifold of hand images embedded in image space---a point which we will return to later. 

With the  class function we can create a model to  generate datasets of images of hands:

\begin{model}
{Find the set of points  $ \mathcal  H=\{ x \in \inputspace  | \, f(x)=1 \},$  and list them (or a specified subset) in some order}. \
 \end{model}
\noindent 
 In general, there will be many  solutions and therefore elements in $  \handspace$, which brings to light the meaning of compression. 
 There will be many such points. For simplicity, assume we can simply list them (in some random order).   Thus, we can use $f(x)$ and a number  (a parameter) to select an element in the list  to unpack the function into an image of a hand.\\
 
 We can also create a generating function from the class function in a similar way: use the class function to generate all images of hands in some order and use that list as the generating function table from integers (parameter space) to images. \\
 
In this way, we can talk about  some types of models (models of stacked datasets) as functions represented/encoded by  neural networks or other machine learning systems. 
The same reasoning can be applied to  recurrent neural network (RNN) trained for some similar task: e.g., once given an input, the trained network will orbit around dynamical attractor the encodes that ``memory'' or classification output. The interesting point is that the attractor (or the output of a subset of nodes mapped to a single output here for binary functions) that the system enters is invariant under a set of transformation of the inputs, exactly as we discussed before.  \\

We note that  in this example,  class   as well as  generating function are primitive recursive, since the input space is finite.

%
%
\begin{figure*}[t!]
\centering
\includegraphics[width=7.5cm]{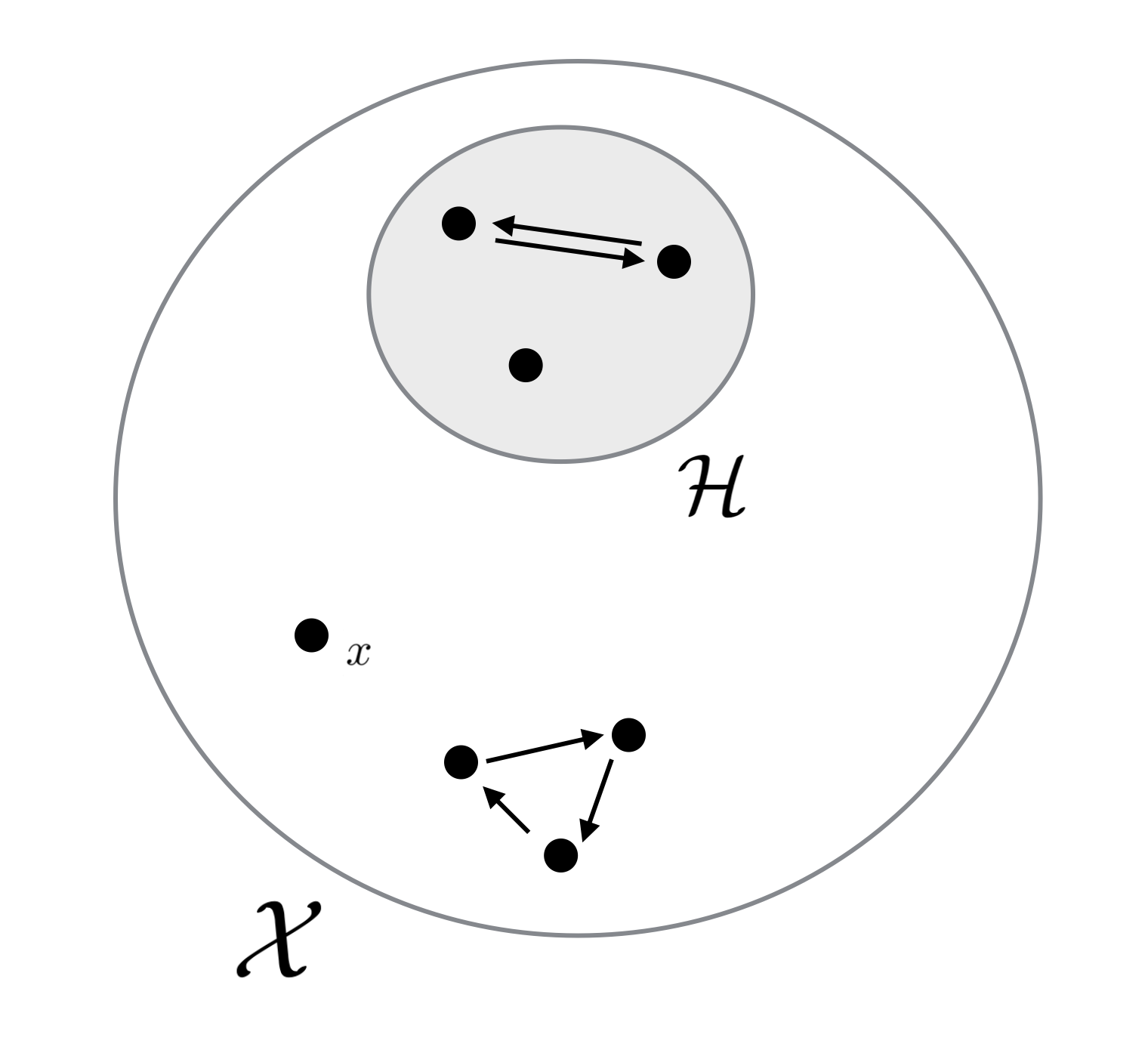}\includegraphics[width=4.8cm]{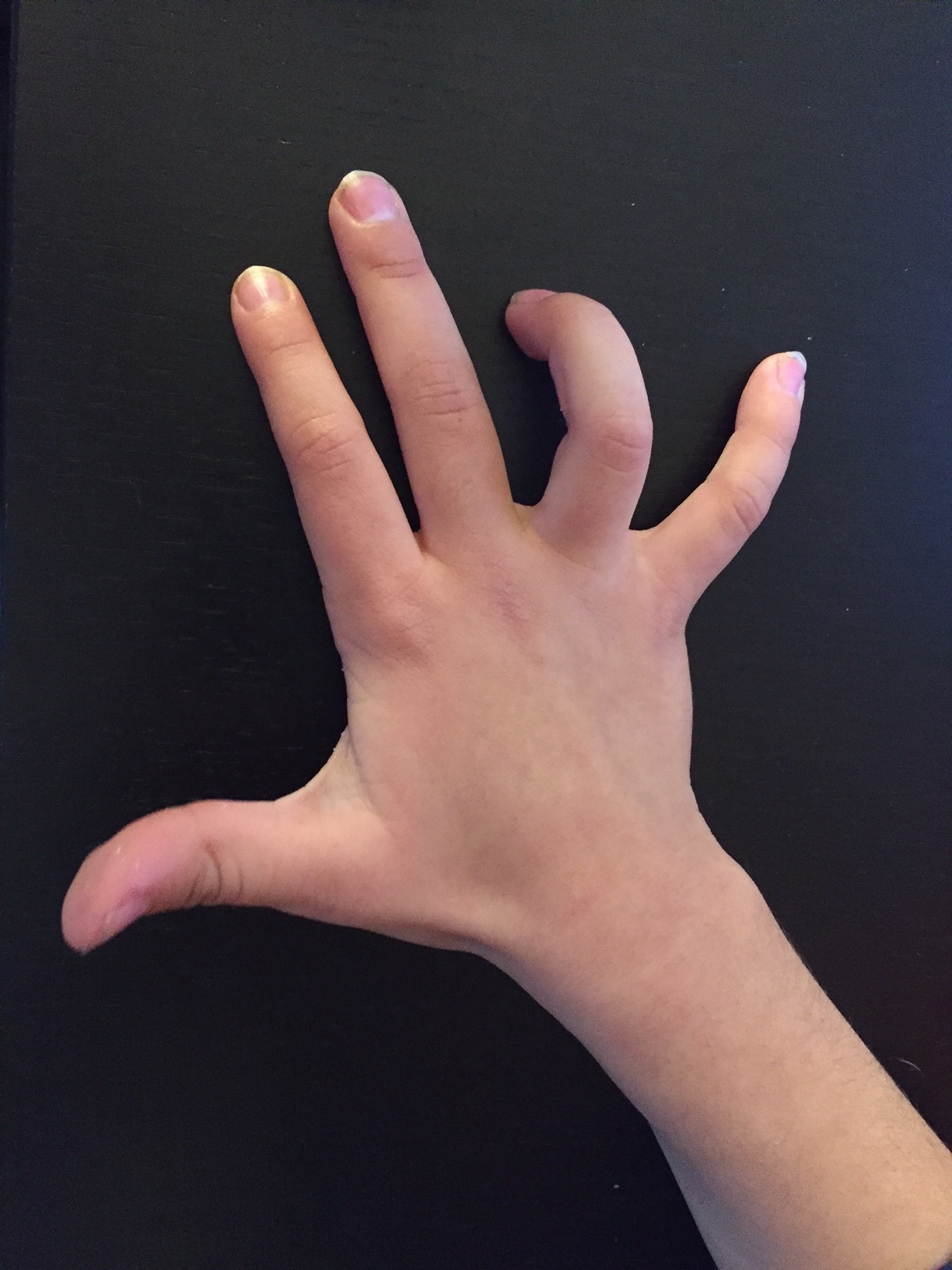}
\caption{Left: Input (image) space $\inputspace$, the subspace of hand images $ \handspace =\{ x \in \inputspace  | \, f(x)=1 \}$ and a example of  automorphism (arrows) leaving the class one set $ \handspace$ invariant.  Right: a sample element $x\in \handspace \subset \inputspace$. \label{fig1}}
\end{figure*}

\section{Algorithmic complexity of a class function}
We have a notion of algorithmic complexity for strings or datasets, the  Kolmogorov complexity, i.e., the length of the shortest  (Turing) program capable of generating the string in some universal language \cite{Cover:2006aa}. We would like to extend this notion  to functions in the present context. This motivates the following definitions. Let an ordering of a countable set $S$ be a function $x_{i}: \mathbb{Z} \rightarrow S$.

\begin{definition} The algorithmic complexity ${\mathcal K}[f]$  of a binary valued function $f(x)$   is the algorithmic complexity of the ordered stacked dataset  $S_{ \mathcal  H}\equiv [ \{x  \in \inputspace | \: f(x)=1\}] = [ \handspace]$ (a string or list) given its ordering,   
 $$
\mathcal K[{f }]= \mathcal K[D_{f } | ordering]   \approx k_{f}^{*} 
 $$ 
 where $k^{*}$ is the length shortest program capable of generating all hand images (the hard part) and then a set of all possible stacks from permutations of the images (easy).
 \end{definition}
 
 Thus, only $k_{f}^{*} $ bits are needed to specify the class function, or, equivalently,  the generating function. 
If we did not know the order, we would need $\log |\handspace| ! \approx |\handspace| \log |\handspace| $ bits to specify it:  {\em  the algorithmic complexity of the dataset is $\mathcal K[{D_{f} }] = k_{f}^{*}  + \log |\handspace| ! \approx  k_{f}^{*}  +  |\handspace| \log |\handspace| $. }
This is large, but much smaller that the length of the dataset, 
$\mathcal K[{D_{f} }]  << l(D_{f})=|\handspace|\, \log |\inputspace|$.  
Ignoring the program length  $k^{*}$  (which we assume is small compared to the other quantities), compression results from us being able to use  $\log |\handspace| $ bits to specify a hand image, instead of the full $ \log |\inputspace| $, and the compression ratio is $\rho=   \log |\handspace | / \log |\inputspace|$.

\section{The group of invariances of the class function}
A  class function defines an equivalence relation (i.e., images in the same class are equivalent). This motivates the following definition (Figure~\ref{fig1}).
\begin{definition} 
 Given a class function $f(x): \inputspace \rightarrow \mathbb B$, we define the set of invariant transformations of  the function (or more simply invariances) to be automorphisms of the domain space $T:\inputspace  \rightarrow \inputspace $ such that   $\forall x \in  \mathcal  H$ we have $T(x) \in  \handspace =\{ x \in \inputspace  | \, f(x)=1 \} $. The set of all such automorphisms forms a group, $\mathcal G_{f}$, the group of invariances of the class function, which acts on input (image) space.
 \end{definition}
Examples of such automorphisms are permutations of the elements of particular class in   $\inputspace$ that leave the other one alone (permutation on $\handspace$ x Identity on $\bar \handspace$). More generally, $\mathcal G_{f}$ contains as elements  all permutations that leave elements in the same class. \\

Also, $\mathcal G_{f}$ is a big set, with as many elements as permutations of elements in $\handspace$ multiplied by those of its complement, 
$$
G_{f} \sim S_{\handspace} \times S_{\bar{\handspace}} 
$$\\

That  $\mathcal G_{f}$ forms a group is easy to see: it has an identity, composition of invariances is an invariance, each  invariance has an inverse, and composition is associative. 
It is immediate that the group of  invariances is equivalent to the class  function as well (up to labeling). 

\begin{theorem} The group  of invariances of a class function $f(x): \inputspace \rightarrow \mathbb B$  uniquely determines the  class function up to labeling.
\end{theorem}

\underline{Proof}: by definition,  $f(x)=f(x')$ if and only if there exists  $T\in \mathcal G_{f}$  such that $T(x) =x'$. Thus, the set of invariances partitions $\inputspace $ into subsets which can be assigned a unique, distinct label (the equivalence class), reproducing the class function.\\

Similar reasoning applies to  an $q$-class (multiclass) functions.  Each class has associated a group of invariances of the set of images.  Thus, learning a class is equivalent to learning the associated function invariance group (up to label, i.e., up to $\log q!$ bits, with $q$ the number of classes).\\

Thus class functions are equivalent to subgroups of invariances or permutations, so by Cayley's theorem, to some group (every group is isomorphic to a group permutations). For instance, the group of rotations acts on image space an appears as a homeomorphism from the rotation group to the group of permutations of the hand image space (a permutation representation of rotations). \\

We can see that in this framework,  {\bf function, invariance group and dataset algorithmic complexities are essentially equivalent notions}.  It is the structure of the group of invariances that is related to the depth of the associated generating function, which can be seen as using the group actions to iterate from image to image. For instance, rotations of images (group action) can sequentially generate many of the images in the stack. This is a ``deep'', recursive operation (there is a Lie group involved), and one that a classifier can exploit.


\section{The efficiency of implementation of a class function by a neural network}
Although neural networks can be seen to implement functions (or models, group of invariances or datasets), they may be inefficient implementations. We can imagine, for instance, a network with  dead-end or even disconnected nodes.  A neural network may be defined, essentially, by a set of connectivity matrices and biases  (we assume the node activation  function to be constant).  This motivates the definition of implementation length. \\

\begin{definition}
  The implementation length of a neural network, $l(\mathcal N)$, is the minimal program length required to specify the architecture and parameter values of $\mathcal N$ (the algorithmic complexity of its architecture). 
  \end{definition}
  
  We note that implementation length is directly related to the number of parameters in the networks's architecture, and hence to the amount of data it will need for training. \\

We can now define the concept of implementation efficiency of a function (or model) by a network, which quantifies how succinct is the implementation of the class function. \\

\begin{definition}Let network $\mathcal N_{f}$  implement a function $f(x)$. The  efficiency of the  network is defined by the ratio of the algorithmic complexity of the function and the minimal description length of the network implementation, which we can assume lies between 0 (inefficient) and 1 (optimal) after some normalization:
$$
 \mathcal E[\mathcal N_{f}] = { {\mathcal K}[f] \over l(\mathcal N_f)} 
 .$$
 \end{definition}

\section{The entropy of a class function; Deep and shallow functions}

Can we define the entropy of a class function?  
 We know that the stacked dataset must contain at least $|\handspace| \log |\handspace|$ bits, and no more than  $|\handspace| \log |\inputspace| = l(D_{f})$.
 We recall here that entropy rate is closely related to Lempel-Ziv  compression \cite{Cover:2006aa}.

\begin{definition}
Let $h_{D_{f}}$ be the entropy rate of the dataset. The total (Shannon) information or entropy of the dataset/model/function is $H[D_{f}]=h_{D_{f}}\, | \handspace|$.
We say  a classification function  is deep when its algorithmic complexity is much smaller than its entropy, i.e.,  $K[D_{f}] <<H[D_{f}] \approx  l_{LZW}[D_{f}] \approx |\handspace| \, h_{D_{f}}$.
\end{definition}
Intuitively, a function is deep if it is doing a lot (generating entropy\footnote{One is tempted to make use here, or a connection to, ideas in the mathematics of fractals, such as fractal dimension, for a metric that relates to this idea of generating entropy or ``filling'' output space.}) despite being algorithmically simple. Consider the following description lengths for the dataset, starting from the initial long one:
\begin{eqnarray*}
d &=& l(D_{f}) = |\handspace| \log |\inputspace| \\
  d_{lzw}&=& l(D_{f}) \, h_{D_{f}} =  h_{D_{f}}\, |\handspace| \log |\inputspace|  \\
  d_{\mathcal K}&=& \mathcal K[D_{f}] = k^{*} +  |\handspace| \log |\handspace|
\end{eqnarray*}
We note that it is known that for most, but not all datasets, $d_{lzw} \approx d_{\mathcal K}$ \cite{Cover:2006aa}.
Examples of deep generating functions from $\mathcal P = \mathbb B ^{m} \rightarrow \inputspace = \mathbb B ^{n}$ are ``For a given ordered input of $m$ bits, move $2^{m} * n$ positions to to the right  in the binary expansion of $\pi$ and extract the following $n$ bits '' using algorithms from Ramanujan's work \cite{Borwein:1989aa}. Or, ``using an elementary Cellular Automaton such as Rule 110 \cite{Wolfram:2002aa} with dimension $n$ and some random initial condition, carry out $2^{m}$ steps and return that line of $n$-bits.''  In both cases, many iterations are needed. The functions are highly recursive (basically performing the same operation over and over), with at least as many recursions as there are ``images'' in the stack ($ |\handspace|$).  The role of recursion in computation is also discussed by Bennett \cite{Bennet:1988aa}, who proposed the concept of logical depth of a string at a significance level $s$ as the minimal number of steps $t$ required to compute it with $s$ bits more than the minimum program length. \\

On the other extreme, we may talk of ``shallow'' functions, which require the direct use of a look-up table to output data. An example would be a function to describe Chaitin's constant  $\Omega$ or other truly incompressible datasets.  As we discuss below, a shallow NN is appropriate for such a task, but deep functions benefit from the availability of the {\tt for-loops} offered by multilayer or recurrent networks.\\

\section{Shallow vs. Deep vs. Recurrent networks}
A neural network can be seen as implementing a program to compute a function as a sequence of steps in a special programming language. In this language,  each layer represents a set of parallel computational steps in a sequence of several possible.  Each step uses the same basic operations.  As we saw, the function thus encoded provides the means for compression of datasets which can then be represented efficiently by the equation $f(x)=1$. \\

A Shallow network is a network of the form \cite{Mhaskar:2016aa}

$$
\mathcal N (x) = \sum_{k=1}^{N} a_{k}\,  \sigma( \langle w_{k}  , x \rangle+ b_{k} ) .
$$

A deep network can be recursively defined by $h^{1}=x$ and  $N (x) = h^{L+1}$, where  
$$
 h^{l+1} =   \sum_{k=1}^{N_{l}} a_{k}^{l} \, \sigma( \langle w_{k}^{l} , h^{l}\rangle + b_{k}^{l} )  
$$
for $l=1,..,L$, which includes a SNN as a special case with only one layer. 

An RNN has the form
$$
h^{t+1} =   \sum_{k=1}^{N} a_{k} \, \sigma( \langle w_{k} , h^{t}\rangle  + \langle v_{k} , x^{t}\rangle + b_{k} )  
$$
where   $N (x^{t}) = \sigma_{o} (\langle w_{o}, h^{t} \rangle +b_{o}).$
\medskip

Feedforward NNs  have been shown to be able to approximate any multivariate primitive recursive function \cite{Cybenko:1989aa, Hornik:1991aa}. Moreover,  if the function to be approximated is compositional 
or recursive,   e.g., 
$$
f(x_1, с с с , x_8) = h_3 \left(h_{21} \left(h_{11} \left(x_1, x_2\right), h_{12}\left(x_3, x_4\right)\right), h_{22}\left(h_{13}\left(x_5, x_6\right), h_{14}\left(x_7, x_8\right)\right) \right)
$$
(which is best visualized as a compositional tree), then it is known that a  hierarchical, compositional  network performs better (will need less training data) \cite{Mhaskar:2016aa}. This is a way of saying that if the data we want the NN to learn is in some sense simple (compositional or recursive), then simpler structure networks (hierarchical) will perform better  (given the same amount of training data) than shallow networks (SNNs), because they can achieve the same accuracy with simpler architectures (fewer nodes or ``complexity'').  \\

Thus, although shallow networks can approximate any function, they are limited  as a programming language in terms of their efficiency. Compositionally (depth) provides NNs more programming power.  \\

Intuitively, a random function can be efficiently encoded by a shallow network, since all that can be done is provide the function table, which is essentially what a SNN will do. \\


If a dataset has low algorithmic complexity (allows for a short description) yet its model/function has high entropy, it is fairly intuitive that generally speaking deep networks will provide more succinct, and hence easier to train network structures than shallow ones, which are limited to one step computations (smaller programming language repertoire) and essentially to producing function tables \cite{Poggio:2016aa}. \\

However, a DNN is again a limited type of system as a programming language: the number of steps it can carry out are fixed once and for all. In essence, a DNN can only carry out a fixed, finite number of {\tt for-loop} steps and can be seen as the equivalent to primitive recursive functions (see, e.g., \cite{Wolfram:2002aa}).  It is not capable of universal computation, as its depth is fixed. Once trained, it computes a primitive recursive function: once its weights are determined, is deterministic w.r.t its inputs.  Recurrent neural networks (RNNs), on the other hand, are known to be Turing complete \cite{Siegelmann:1995aa,Goodfellow:2016aa}. Recurrence  enables universal modeling.  In an RNN, outputs depend on inputs and its initial state. Thus, it can respond to the same input in different ways, depending on its state (program).  In computation theory terms, it includes the analog of the  {\tt while} loop, which extends their reach and makes them universal. This gives it a much larger repertoire, including memory, and puts it the class of $\mu$-recursive functions or universal Turing machines. 
\\

Thus, although all NNs are universal with regard to primitive recursive functions, RNNs, then DNNs and last SNNs should form an efficient encoding hierarchy and, hence, a hierarchy in terms of training set requirements (since the architecture of the network determines how many  trainable parameters it has).
 We formalize this in the following conjecture, which is in the same spirit as the one in \cite{Mhaskar:2016aa}:

\begin{conjecture}[Network Hierarchy]
The implementation efficiency of deep functions (with $\mathcal K[f] <<  \handspace[f]$) by networks is highest for recurrent,  then  deep, then shallow networks, i.e., 
$$
\mathcal E[\mathcal N^{recurrent}_{f}]   \ge   \mathcal E[\mathcal N^{deep}_{f}]  \ge \mathcal E[\mathcal N^{shallow}_{f}] 
$$
Furthermore, the inequalities are stronger the larger the ratio of  entropy to algorithmic complexity of the function, $ \mathcal R_{f}=  \handspace[f] / \mathcal K[f] $.
\end{conjecture}

We sketch a possible  proof:
\begin{enumerate}
\item
SNNs and DNNs  are equivalent computationally to PR functions, a subset of the language used in $\mu$-recursive functions. RNNs are universal. This is already known \cite{Siegelmann:1995aa}. 
\item
Although we deal here with functions that are computable by SNNs (using lookup tables, since the input and output space is finite), access restricted to a subset of a programming language can only lead to longer algorithmic descriptions and therefore longer implementation lengths and heavier architectures (more parameters to train). This is immediate as well. In particular, not having access to {\tt for-loops} (SNNs), or only to a fixed finite number of them (as in DNNs) is a disadvantage.
\item We can take the viewpoint of the generating function angle (equivalently to class function or model). We have $|\handspace| \log |\handspace|$ bits of information as inputs (generating function parameters) to expand into the dataset, which has $|\handspace| \log |\inputspace|$ random looking bits. This is to be achieved by iteration (repeated calculation). In order to generate randomly looking sequences, recursion, reuse of prior calculations must be employed ({\tt for-loop} steps)---copying inputs many times will not generate entropy.
\end{enumerate}

Deep functions will benefit greatly from DNN architectures. A shallow function computing an algorithmically random string will require a shallow network with essentially as many nodes as the size of $|\inputspace|$ (to function as  a lookup table). In the case of a deep (simple) function, a DNN or an RNN, depending on how deep the function is, will require fewer nodes (and training points), because this architecture can exploit and represent the simplicity of the underlying function.  Paralleling the reasoning in \cite{Mhaskar:2016aa}, in this case we would need $\log |\inputspace|$ nodes (one per input), but it would be interesting to getter a better estimate on the number of nodes  as a function of the size of both input and hand-space, and the algorithmic complexity of the model. 

\section{Modeling in the Solomonoff prior}

With regard to the practical implication of this conjecture,  we can extend it  making reference to the Solomonoff prior\footnote{ Alternatively, we can argue the relevance of entropic strings from a anthropic algorithmic principle: cognitive system can one model deep functions.} \cite{Cover:2006aa}. We recall here the  (Solomonoff)  algorithmic or universal (un-normalized) probability $P_\U(x)$ of a string $x$. This is the probability that a given string $x$ could be generated by a random program. An important result is that this is given by 
$ P_\U(x)
\approx 2^{-K_\U(x)}$  \cite{Li:2008aa}.
Thus,  short programs contribute most to the probability of observing a given data string and, furthermore,   the probability of a given string to be produced by a random program is dominated by its Kolmogorov complexity.  We can hypothesize that such is the case in the real world: most datasets, models or functions in the real world will be short but entropic (simple programs are know to produce entropic datasets \cite{Wolfram:2002aa}).

\begin{conjecture}
The implementation efficiency of networks is higher for deep than shallow networks in the Solomonoff universal prior. That is, for most dataset derived functions (with high universal probability), 
$$
\mathcal E[\mathcal N^{r}_{f}]   \ge   \mathcal E[\mathcal N^{d}_{f}]  \ge \mathcal E[\mathcal N^{s}_{f}] 
$$
with the probability function defined by the Solomonoff prior.
Furthermore, the inequalities are stronger the larger the ratio of  entropy to algorithmic complexity of the function, $ \mathcal R_{f}=  \handspace[f] / \mathcal K[f] $.
\end{conjecture}

The proof rests on showing that most simple functions are also  entropic.

\section{Perturbation of an optimal,  well trained    network}

Here we wish to consider a scenario in which a) one class is much smaller than the other (l$|\handspace|/|\inputspace| <<1$) and b) the smaller class is very large still compared to the class function program length (the function is deep). \\

For instance, let us consider the simplest  trained NN with perfect performance in identifying hand images. To train this NN
, suppose that we have used a  short, simple program  to generate  a large number of  hand images by transformation of a template. For simplicity, let us consider only 3D translations and rotations of the hand. Those involve 6 degrees of freedom. Allowing for 64 bits per parameter, we have some $2^{64\cdot 6} \sim 10^{116}$ possible  hand images, to which we need to add backgrounds using a random number generator (another big space). This may seem big, but it is small compared to the input space  $2^{24\cdot 10^{6}}$ in monochrome 24 bit images.  \\ 
\\

{\bf Perturbation of  inputs to the NN:} Let us suppose, then, that we have found a perfectly performing, efficient network  $D$ for our binary classification problem. What can we say about its connectivity? For comparison, let $D_{r}$ be a network with the same architecture  but random weights. 
If we input an arbitrary image of a hand $h \in \inputspace$ to $D$ it will certainly output 1 (for hand, $D(h)=1$, while the outcome with the second network  will be random. Now suppose that we  perturb the hand image. Since the space of hand images is much smaller than the space of not-hands (low relative entropy of $f(x)$), the effect on $D$ will be to change its output to 0 with high probability, 
$D(h+\delta h)=0$, while the effect on the random network will be random. We can summarize this by averaging over the space of perturbations, 
$$
\langle D(h) - D(h+\delta) \rangle =1, \:\: \:\: \langle D_{r}(h) - D_{r}(h+\delta) \rangle =0
$$
Moreover, if we ask the same question with ``not-hand'' input images $\bar h \in \inputspace$ , a perturbation will not have much effect in either case. In the first case, this is simply because the subspace of not-hand images is much bigger than the subspace of hands. A perturbation of ``not-hand'' will, with probability near one, leave the image in the same class. In the case of the random network, the outcome will be random, and the change again average out.
$$
\langle D(\bar h) - D(\bar h+\delta) \rangle =0, \:\: \:\:  \langle D_{r}(\bar h) - D_{r}(\bar h+\delta) \rangle =0
$$
So we have two similarly sized NNs, and their response to perturbation is very different depending on the class of the image they are processing. \\

We can summarize this in the following theorem:
\begin{theorem} In an efficient, well trained NN encoding a  class function in which one class is much smaller than the other,  a perturbation of the NN input when fed by an example in the smaller class will propagate efficiently with high probability, but not in a randomly parameterized NN with the same architecture or for activations in the other, larger class.
\end{theorem}

{\bf Perturbation of nodes in  the NN:}
What about perturbations of a  node in the network? An efficient but shallow NN implementing a deep function with a large range of size $|\inputspace|$ will use many nodes, basically as many as the size of $|\inputspace|$ (to see this one can refer to the work in \cite{Mhaskar:2016aa}, with $\epsilon$ seen as scale parameter to discretize the input space into hypercubes of size $\epsilon^{1/m}$). Perturbation of the activation function or value  of a node will only affect the output for some input values, but not all the input space.  For an efficient DNN, on the other hand, each node will have to play a crucial role in the calculation of {\em all} the values in the function table. A perturbation of the node will have a big impact on the entire function.  The impact on a performing deep efficient network will again be to take to the larger class. If it did not, we could simply remove that node and get an equally performing simpler network.  Perturbations will always lead to classification into the larger class space.

\begin{theorem} In an efficient, well trained NN encoding a  class function in which one class is much smaller than the other,  a perturbation of a node  NN  when fed by an example in the smaller class will propagate efficiently with high probability if the network is deep, but not in a shallow NN, or randomly parameterized NN with the same architecture or for activations in the other, larger class.
\end{theorem}

This potentially establishes a  link between the {\em perturbation complexity index (PCI)} measured by perturbing human cortical neuronal networks by TMS   \cite{Casali:2013aa} and algorithmic complexity. 

Furthermore, perturbations in such a network will appear to be decorrelated at different locations due to the non-linear nature of signal transmission in NNs. This provides the ``information'' aspect and a potential explanation for hard to compress multichannel EEG streams using LZW in the PCI \cite{Casali:2013aa}.

Finally, we note that although hierarchical networks represent a subset of  all multivariate functions \cite{Mhaskar:2016aa}, it would appear that RNNs can bypass this limitation, as they are universal, by affording compositional DNNs of infinite depth.


\section{Discussion}
What mechanisms are in place to evolve simple programs or efficient networks?   A network is actually a computer program, and it may be a good or a  poor implementation of the model, so we may ask how long or even how compressible this program is as a string. Such program may contain much underused,  inefficient or even irrelevant code, while maintaining great classification performance and in fact implementing a simple model. For example, the program, in general Turing machine terms,  could have lengthy code to write and erase a billion random digits before outputting the class label. In a network we could have orphaned nodes that don't contribute to the final computation. So the program or network architecture, as a string, may actually be huge and  incompressible (algorithmically very complex).  

How such an inefficient program has come to be found during training among all possible others of that size is an interesting question.  We can think of this aspect of the problem from the point of view of evolutionary programming. Let us suppose we set up a problem with a simple program as the solution.  We can imagine we have setup an evolutionary system where programs are bred to perform well on the task, starting from small programs that aggregate to larger ones at each generation. It is fairly clear that in such an organic-growth type of  search, the large, inefficient but performing program will never be found, since smaller solutions are  at hand, and the space of programs becomes huge quickly with program size. From this growth structure, hierarchies and power laws will emerge naturally \cite{Ravasz:2003aa}. Since we  may know (because we defined it) the underlying model to be very simple, searching for solutions from short to longer  programs will speed up the process significantly.

We can conjecture that as real neural networks develop in brains during learning, they must implement simplicity principles in the process, such as sparsity.  One mechanism is a synaptic cap growth model, which seems to force in a natural way the emergence of sparse networks \cite{Hahnloser:2002aa}. Of course, sparsity is one way to approach the $\mathcal K$  minimum. Another example is \cite{Brunel:2016aa}, where it is shown that one way to create robust networks with high memory capacity (attractors or learned sequences) is by a)  sparsity, b) more numerous, stronger bidirectional connections than random networks. In machine learning, e.g., from statistical learning theory to Echo State Networks, sparsity is important and normally used.  Robustness of such networks is also related to good generalization \cite{Bousquet:2002aa} and hence sparsity or, probably more generally, Kolmogorov simplicity.  And in terms of resources, simplicity is great: less memory and computation will be needed.  Scarcity, competition are both helpful in this context, since they will lead to some form of simplicity. This may be the reason why evolutionary search is successful. It provides a gradient towards simple programs. Memory, power/energy and computation time as limited resources will undoubtedly be important for model selection in competitive worlds.\\
 
Why is the universe simple? Is the answer anthropic --- i.e., ``If it wasn't simple, there would not be a discussion'' --- or can we do better?  Regardless of the answer, assuming there is simplicity exists is a valid starting point (as generations of physicists can attest). \\

{\bf Acknowledgements:} This work partially supported by the FET Open Luminous project (this project has received funding from the European Union's Horizon 2020 research and innovation programme H2020-FETOPEN-2014-2015-RIA under agreement No. 686764). The author is very grateful to Tomaso Poggio for his lectures (9.520 Statistical Learning Theory and Applications - MIT) and inspiring discussions on the topics of network/function compositionally, depth and their links to simplicity. 
\footnotesize
\bibliographystyle{plain} 
\bibliography{../bibliographies/kolmogorov}  

\begin{thebibliography}{10}

\bibitem{Barbour:1999aa}
Julian Barbour.
\newblock {\em The end of time}.
\newblock Oxford University Press, 1999.

\bibitem{Bennet:1988aa}
CH~Bennet.
\newblock Logical depth and physical complexity.
\newblock In Rolf Herken, editor, {\em The Universal Turing Machine--a
  Half-Century Survey}, pages 227--257. Oxford University Press, 1988.

\bibitem{Borwein:1989aa}
J.~M. Borwein, P.~B. Borwein, and D.~H~. Bailey.
\newblock Ramanujan, modular equations, and approximations to pi, or how to
  compute one billion digits of pi.
\newblock {\em Amer. Math. Monthly}, pages 201--219, 1989.

\bibitem{Bousquet:2002aa}
Olivier Bousquet and Andre Elisseef.
\newblock Stability and generalization.
\newblock {\em Journal of Machine Learning Research}, 2:499--526, 2002.

\bibitem{Brunel:2016aa}
Nicolas Brunel.
\newblock Is cortical connectivity optimized for storing information?
\newblock {\em Nature}, 19(5), May 2016.

\bibitem{Casali:2013aa}
Adenauer~G. Casali, Olivia Gosseries, Mario Rosanova, M\'elanie Boly, Simone
  Sarasso, Karina~R. Casali, Silvia Casarotto, Marie-Aur\'elie Bruno, Steven
  Laureys, Giulio Tononi, and Marcello Massimini.
\newblock A theoretically based index of consciousness independent of sensory
  processing and behavior.
\newblock {\em Sci Transl Med}, 5(198), 2013.

\bibitem{Cover:2006aa}
Thomas~M. Cover and Joy~A. Thomas.
\newblock {\em Elements of information theory}.
\newblock John Wiley \& sons, 2 edition, 2006.

\bibitem{Cybenko:1989aa}
G~Cybenko.
\newblock Approximations by superpositions of sigmoidal functions.
\newblock {\em Mathematics of Control, Signals, and Systems}, 2(4):303--314,
  1989.

\bibitem{Goodfellow:2016aa}
Ian Goodfellow, Yoshua Bengio, and Aaron Courville.
\newblock {\em Deep Learning}.
\newblock MIT Press, http://www.deeplearningbook.org, 2016.

\bibitem{Hahnloser:2002aa}
Richard H.~R. Hahnloser, Alexay~A. Kozhevnikov, and Michale~S. Fee.
\newblock An ultra-sparse code underlies the generation of neural sequences in
  a songbird.
\newblock {\em Nature}, 419, 2002.

\bibitem{Hofstadter:1979aa}
Douglas Hofstadter.
\newblock {\em Godel, Escher, Bach}.
\newblock Basic Books, 1979.

\bibitem{Hornik:1991aa}
Kurt Hornik.
\newblock Approximation capabilities of multilayer feedforward networks.
\newblock {\em Neural Networks,}, 4(2):251--257, 1991.

\bibitem{Li:2008aa}
Ming Li and Paul Vitanyi.
\newblock {\em An Introduction to {Kolmogorov Complexity} and Its
  Applications}.
\newblock Spriger Verlag, 2008.

\bibitem{Mhaskar:2016aa}
Hrushikesh Mhaskar, Qianli Liao, and Tomaso Poggio.
\newblock Learning functions: When is deep better than shallow.
\newblock Technical Report CBMM Memo No. 045, CBBM, 2016.

\bibitem{Poggio:2016aa}
T~Poggio, Hrushikesh Mhaskar, Lorenzo Rosasco, Brando Miranda, and Qianli Liao.
\newblock Why and when can deep -- but not shallow --networks avoid the curse
  of dimensionality: a review.
\newblock {\em CBMM Memo}, (058), 2016.

\bibitem{Ravasz:2003aa}
Erzsebet Ravasz and Albert-Laszlo Barabasi.
\newblock Hierarchical organization in complex networks.
\newblock {\em Phys. Rev.}, E(67), 2003.

\bibitem{Ruffini:2016ac}
G.~Ruffini.
\newblock An algorithmic information theory of consciousness.
\newblock {\em Submitted to the Neuroscience of Consciousness}, August 2016.

\bibitem{Siegelmann:1995aa}
H.~T. Siegelmann and E.D. Sontag.
\newblock On the computational power of neural nets.
\newblock {\em Journal of computer and system sciences}, 50(1):132--150, 1995.

\bibitem{Wolfram:2002aa}
Stephen Wolfram.
\newblock {\em A New Kind of Science}.
\newblock Wolfram Media, 2002.

\end{thebibliography}

\end{document}